\documentclass{article}
\usepackage{spconf,amsmath,graphicx}
\usepackage{booktabs}
\DeclareGraphicsExtensions{.pdf, .png, .jpg}
\usepackage{makecell}
\usepackage{subcaption}

\title{Extrapolative-Interpolative Cycle-Consistency Learning \\ for Video Frame Extrapolation}
\name{Sangjin Lee\sthanks{E-mail: pandatimo@yonsei.ac.kr}, Hyeongmin Lee, Taeoh Kim and Sangyoun Lee\sthanks{Corresponding Author, E-mail: syleee@yonsei.ac.kr}}
\address{School of Electrical and Electronic Engineering, Yonsei University, Seoul, Korea}

\begin{document}
	%
	\maketitle
	\begin{abstract}
		Video frame extrapolation is a task to predict future frames when the past frames are given. Unlike previous studies that usually have been focused on the design of modules or
		construction of networks, we propose a novel Extrapolative-Interpolative Cycle (EIC) loss using pre-trained frame interpolation module to improve extrapolation performance.
		Cycle-consistency loss has been used for stable prediction between two function spaces in
		many visual tasks. We formulate this cycle-consistency using two mapping functions; frame extrapolation and interpolation. Since it is easier to predict intermediate frames than to predict future frames in terms of the object occlusion and motion uncertainty, interpolation module can give guidance signal effectively for training the extrapolation function.
		EIC loss can be applied to any existing extrapolation algorithms and guarantee consistent prediction in the short future as well as long future frames. Experimental results show that simply adding EIC loss to the existing baseline increases extrapolation performance on both UCF101 \cite{journals/corr/abs-1212-0402} and KITTI \cite{Geiger2013IJRR} datasets.
	\end{abstract}

	\section{Introduction}
	\label{sec:intro}
	
	Video frame extrapolation is a task that predicts future frames, which is very challenging because it requires comprehensive understanding of the objects and motions. It can be used as a key component for video application such as future forecasting, action recognition, and video compression.
	
	Two of the most important problems in frame extrapolation are blurry output image and vulnerability to occlusion. Therefore, recent extrapolation studies have made various efforts to solve these problems through various methods. The extrapolative methods can be classified into three categories: the pixel-based, flow-based, and hybrid methods. Research approached using the pixel-based method \cite{byeon2018contextvp, lu2017flexible, oliu2018folded, lotter2016deep} uses 3D Convolutional Network or ConvLSTM \cite{xingjian2015convolutional} / GRU \cite{cho2014learning} to predict directly future pixels, and expect to learn flow information between frames implicitly while training. However, despite using Generative Adversarial Network (GAN) \cite{goodfellow2014generative} or Variational Autoencoder (VAE) \cite{kingma2013auto}, the pixel-based methods make the image blur due to the recurrent modeling. Another research, approached with the flow-based method~\cite{liu2017video, li2018flow, gao2019disentangling}, obtains target frame by warping the input frame with the predicted target flow map. In contrast to the pixel-based method, sharper image can be obtained by moving the pixels through predicted flows. However, it is vulnerable to objects which have large movements or with occlusions. To solve these problems, some research \cite{villegas2017decomposing, liang2017dual} uses both methods to form a network for complementary purposes which shows better performance. We use this hybrid method as our baseline.
	
	\begin{figure}[t]
		\setlength{\belowcaptionskip}{-15pt}
		\centering
		\includegraphics[width=\columnwidth]{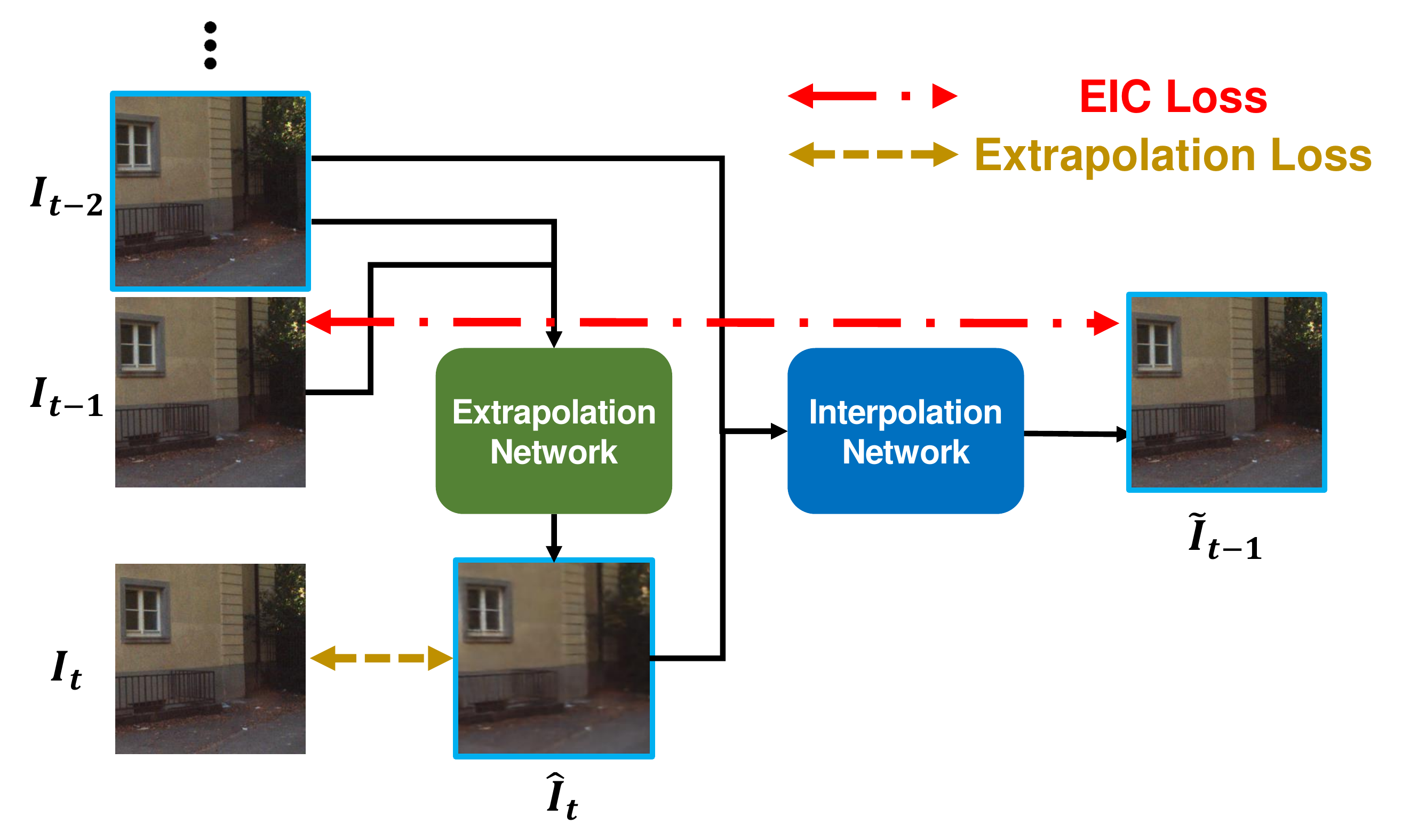}
		\caption{$I_t$ is a target frame and $I_{t-k:t-1}$ are input frames for extrapolation network. The result of extrapolation network~($\hat{I}_t$) and $I_{t-2}$ frame are passed into the interpolation network. By equation (\ref{eic_loss}), we can obtain EIC loss like red dotted line from the result of interpolation network ($\tilde{I}_{t-1}$).}
		\label{main idea}
	\end{figure}
	
	In our research, we look for ways to increase performance by working on designing loss rather than network architecture. Cycle-consistency loss has been used in existing research \cite{CycleGAN2017, meister2018unflow, Kwon_2019_CVPR} and has resulted in improved learning stability and performance. Unlike the research~\cite{CycleGAN2017, meister2018unflow, Kwon_2019_CVPR}, it is difficult to apply cycle-consistency loss to video frame extrapolation in traditional ways, due to the uncertainty of extrapolation. In \cite{meister2018unflow}, they suggests that optical flow performance can be increased by training via forward-backward flow learning. Because optical flow estimation usually requires two frames, this forward-backward consistency is effectively operated. In video extrapolation, \cite{Kwon_2019_CVPR} trains prediction models with forward-backward consistency loss to predict back the past frames. However, training single prediction function from scratch with cycle loss can cause unsuitability in training. In \cite{CycleGAN2017}, they proposes cycle-consistency loss between two mapping functions from scratch for unsupervised image translation. Different from these models, we build new cycle-consistency from extrapolation and interpolation. Because frame interpolation task is easier than extrapolation in terms of the object occlusions and motion uncertainties, it can give stable guidance signal to train the extrapolation function. Specifically, in the case of frame interpolation, all the information to estimate intermediate frame can be obtained at least one adjacent frame. For this reason, our goal is to increase extrapolation performance from the interpolation module by adding cycle-consistency loss.
	
	In this paper, we propose a loss simply applicable to frame extrapolation called Extrapolative-Interpolative Cycle (EIC) loss. Then, we add it to existing extrapolation loss to identify performance differences. To check the effectiveness on the hard predictions, we verify the results not only in the short future but also in long future frames.   Results show that our novel loss gives huge performance increments in various settings.
	
	\section{Proposed Algorithm}
	\label{sec:majhead}
	
	\subsection{Overview}
	\label{ssec:overview}
	
	The overall concept of Extrapolative-Interpolative Cycle (EIC) loss is described in Fig. \ref{main idea}. The system for EIC loss consists of two functions; extrapolation network and interpolation network. Note that any kinds of algorithm can be used for the two functions if equations (\ref{extra_function}) and (\ref{inter_function}) are satisfied. For frame extrapolation network $f_{e}$, we define the function which predicts the next future frame $\hat{I}_{t}$ as
	
	\begin{equation}
	\label{extra_function}      
	\hat{I}_{t} = f_{e}(I_{t-1}, I_{t-2}, ..., I_{t-k})
	\end{equation}
	
	\noindent where $I_{t-k:t-1}$ indicates the given frames, $k$ is the number of given past frames and $I_t$ is the target frame. For the frame interpolation network $f_{i}$, we define the function which predicts the intermediate frame $\tilde{I}_{t-1}$ as
	
	\begin{equation}
	\label{inter_function}      
	\tilde{I}_{t-1} = f_{i}(I_{t-2}, I_{t})
	\end{equation}

	After predicting the next frame $\hat{I}_{t}$, since the interpolation network generates an intermediate frame when two consecutive frames are given, we can re-synthesize the ($t-1$)-th frame from the ($t-2$)-th and ($t$)-th frames.
	
	\begin{table}[ht!]
		\setlength{\belowcaptionskip}{-0pt}
		\centering\renewcommand\arraystretch{1}\resizebox{\columnwidth}{!}{
			\begin{tabular}{cccc}
				& DVF \cite{liu2017video} & SuperSlomo \cite{jiang2018super} & SepConv \cite{niklaus2017video}\\
				\midrule
				PSNR & 34.31 & 33.92 & 34.38\\
				SSIM & 0.949 & 0.949 & 0.951\\
				\bottomrule
		\end{tabular}}
		\caption{Performance (PSNR and SSIM \cite{wang2004image}) of frame interpolation modules on UCF101.}
		\label{interpolation modules}
	\end{table}%
	
	\begin{table}[ht!]
		\setlength{\belowcaptionskip}{-10pt}
		\centering\renewcommand\arraystretch{1.3}\resizebox{\columnwidth}{!}{
			\begin{tabular}{c|cc|cc}
				& \multicolumn{2}{c|}{UCF101} &\multicolumn{2}{c}{KITTI}\\
				& PSNR & SSIM & PSNR & SSIM\\
				\hline
				Baseline & 27.10 & 0.861 & 22.61 & 0.730\\
				\hline
				+ DVF ($\lambda=1$) & 28.24 & 0.862 & \textbf{23.33} & 0.760\\
				+ SuperSlomo ($\lambda=1$) & 28.01 & 0.868 & 22.86 & \textbf{0.759}\\
				+ SepConv ($\lambda=1$) & 28.20 & 0.876 & 22.88 & 0.761\\
				\hline
				+ DVF ($\lambda=0.1$)  & \textbf{28.34} & \textbf{0.877} & 23.17 & \textbf{0.771}\\
				+ SuperSlomo ($\lambda=0.1$) & \textbf{28.10} & \textbf{0.883} & \textbf{22.98} & 0.722\\
				+ SepConv ($\lambda=0.1$) & \textbf{28.29} & \textbf{0.889} & \textbf{23.01} & \textbf{0.773}\\
				\hline
				+ DVF ($\lambda=0.01$) & 28.12 & 0.863 & 23.05 & 0.761\\
				+ SuperSlomo ($\lambda=0.01$) & 27.96 & 0.865 & 22.90 & 0.731\\
				+ SepConv ($\lambda=0.01$) & 28.09 & 0.873 & 22.95 & 0.765\\
				\hline
		\end{tabular}}
		\caption{Performance (PSNR and SSIM) of video frame extrapolation on UCF101 and KITTI with and without our EIC loss.}
		\label{main results}
	\end{table}
	
	\begin{table*}[ht!]
		\setlength{\belowcaptionskip}{-15pt}
		\centering\renewcommand\arraystretch{2}\resizebox{\linewidth}{!}{
			\begin{tabular}{c|cccc|cccc}
				\toprule
				\hline
				& \multicolumn{4}{c|}{\large{UCF101}} & \multicolumn{4}{c}{\large{KITTI}}\\
				\hline
				\large{Frame} & \large{$\hat{I}_{t}$} & \large{$\hat{I}_{t+1}$} & \large{$\hat{I}_{t+2}$} & \large{$\hat{I}_{t+3}$} & \large{$\hat{I}_{t}$} & \large{$\hat{I}_{t+1}$} & \large{$\hat{I}_{t+2}$} & \large{$\hat{I}_{t+3}$} \\
				\hline
				\large{Baseline} & 27.10 / 0.861 & 22.45 / 0.770 & 19.44 / 0.688 & 17.26 / 0.613 & 22.61 / 0.730 & 19.58 / 0.638 & 17.32 / 0.563 & 15.61 / 0.510\\
				\hline
				\large{+ DVF} & \makecell{28.24 / 0.862 \\ \textbf{(1.14) / (0.001)}} & \makecell{24.28 / 0.771 \\ \textbf{(1.83) / (0.001)}} & \makecell{21.85 / 0.691 \\ \textbf{(2.41) / (0.003)}} & \makecell{20.13 / 0.635 \\ \textbf{(2.87) / (0.022)}} & \makecell{23.33 / 0.760 \\ \textbf{(0.72) / (0.030)}} & \makecell{20.86 / 0.665 \\ \textbf{(1.28) / (0.027)}} & \makecell{18.88 / 0.593 \\ \textbf{(1.56) / (0.030)}} & \makecell{17.36 / 0.537 \\ \textbf{(1.75) / (0.027)}}\\
				\large{+ SuperSlomo} & \makecell{28.01 / 0.868 \\ \textbf{(0.91) / (0.007)}} & \makecell{24.20 / 0.785\\ \textbf{(1.75) / (0.015)}} & \makecell{21.87 / 0.720\\ \textbf{(2.43) / (0.032)}} & \makecell{20.22 / 0.670\\ \textbf{(2.96) / (0.057)}} & \makecell{22.86 / 0.759 \\ \textbf{(0.25) / (0.029)}} & \makecell{20.25 / 0.666 \\ \textbf{(0.67) / (0.028)}} & \makecell{18.32 / 0.599\\ \textbf{(1.00) / (0.036)}} & \makecell{16.94 / 0.550\\ \textbf{(1.33) / (0.040)}}\\
				\large{+ SepConv} & \makecell{28.20 / 0.876 \\ \textbf{(1.10) / (0.015)}} & \makecell{24.17 / 0.789 \\ \textbf{(1.72) / (0.019)}} & \makecell{21.51 / 0.722 \\ \textbf{(2.07) / (0.034)}} & \makecell{19.50 / 0.666\\ \textbf{(2.24) / (0.053)}} & \makecell{22.88 / 0.761 \\ \textbf{(0.27) / (0.031)}} & \makecell{20.30 / 0.673\\ \textbf{(0.72) / (0.035)}} & \makecell{18.26 / 0.600\\ \textbf{(0.94) / (0.037)}} & \makecell{16.70 / 0.541\\ \textbf{(1.09) / (0.031)}}\\
				\hline
				\bottomrule
		\end{tabular}}%
		\caption{Performance (PSNR / SSIM) of long future frame extrapolation on UCF101 and KITTI. The values in parenthesis are the relative performance improvements between baseline and additional loss. Four frames ($\hat{I}_{t}$, $\hat{I}_{t+1}$, $\hat{I}_{t+2}$ and $\hat{I}_{t+3}$) are predicted.}
		\label{longterm results}
	\end{table*}%
	
	\subsection{Learning with EIC Loss}
	\label{ssec:learning with EIC Loss}
	
	To train the extrapolation network, various combinations of losses have been used in previous works (Yellow dased line in Fig. \ref{main idea}). We can formulate these losses as
	
	\begin{equation}
	\label{extra_loss}      
	\mathcal{L}_{extra} = \mathcal{L}_{E}(\hat{I}_{t}, I_{t}) + \alpha\mathcal{L}_{G}(\hat{I}_{t}, I_{t}) + \beta\mathcal{L}_{R}
	\end{equation}
	
	\noindent where $\mathcal{L}_{E}$, $\mathcal{L}_{G}$ and $\mathcal{L}_{R}$ indicate error-based, generative model-based and regularization loss, respectively. The choice of the $\alpha$, $\beta$, $\mathcal{L}_{E}$, $\mathcal{L}_{G}$ and $\mathcal{L}_{R}$ is algorithm-specific. Error-based loss ($\mathcal{L}_{E}$) can be either $\mathcal{L}_1$ loss or $\mathcal{L}_2$ loss.
	Examples of generative model-based loss ($\mathcal{L}_{G}$) are GAN loss or KL-divergence in the VAE loss. Smoothing loss such as total variation (TV) in the pixel or flow domain can be used as regularization loss ($\mathcal{L}_{R}$).
	
	Our proposed Extrapolative-Interpolative Cycle (EIC) loss can simply be added to the existing loss which is defined as
	\begin{equation}
	\label{eic_function}
	\tilde{I}_{t-1} = f_{i}(I_{t-2}, \hat{I}_{t})
	\end{equation}
	\begin{equation}
	\label{eic_loss}
	\mathcal{L}_{EIC} = \mathcal{L}_1(\tilde{I}_{t-1}, I_{t-1})
	\end{equation}
	
	\noindent where $\tilde{I}_{t-1}$ is obtained by entering $\hat{I}_t$ as the input instead of $I_t$ in equation (\ref{inter_function}) (Red dashed line in Fig. \ref{main idea}). We use pre-trained interpolation networks for $f_{i}$ without fine-tuning.
	Finally, our total loss to train the extrapolation network can be summarized as
	
	\begin{equation}
	\label{total_loss}
	\mathcal{L}_{total} = \mathcal{L}_{extra} + \lambda\mathcal{L}_{EIC}
	\end{equation}
	
	\noindent where $\lambda$ is a hyper-parameter which balances the extrapolation loss and cycle guidance loss. We report the performance differences depending on the choice of $\lambda$ in the Section \ref{sec:experiments}. The network can be trained end-to-end with our loss and almost similar training time is required. In the test phase, only the extrapolation network is used because our loss with the interpolation network is only used in the training phase. Therefore, proposed method including our model does not require additional computations and parameters at the test phase compare to existing methods.

	\begin{figure*}[ht!]
		\setlength{\belowcaptionskip}{-20pt}
		\includegraphics[width=\textwidth]{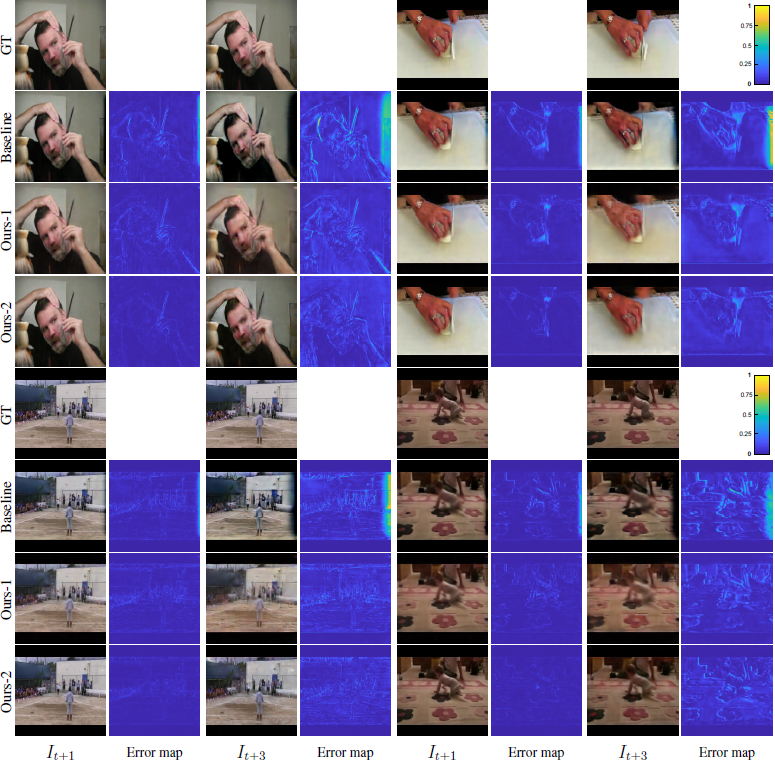}
		\vspace{-5mm}
		\caption{Qualitative results of long future frame extrapolation on UCF101 dataset, we denote Ours-1 and Ours-2 for DVF and SepConv, respectively. Error maps represent errors between GT images and each predicted images.}
		\vspace{1mm}
		\label{fig:vis_results}
	\end{figure*}

	\section{Experiments}
	\label{sec:experiments}
	
	\subsection{Settings}
	\label{ssec:settings}
	
	We select Dual Motion GAN \cite{liang2017dual} as our baseline since it is hybrid method and representative algorithm for frame extrapolation. Different from Dual Motion GAN, six previous frames ($k=6$) are given as input for video extrapolation. Vimeo90K \cite{xue2019video} dataset is used for training and UCF101 \cite{journals/corr/abs-1212-0402} and KITTI \cite{Geiger2013IJRR} datasets are used for evaluation. Other settings such as loss function ($\mathcal{L}_{extra}$) and learning rate are identical with the prior research, and our EIC loss is added using equation (\ref{total_loss}).
	
	For EIC loss, three representative pre-trained video frame interpolation models are used in our research: Deep Voxel Flow (DVF) \cite{liu2017video}, SuperSlomo \cite{jiang2018super} and SepConv \cite{niklaus2017video}. We report the interpolation performance of these models on the UCF101 dataset in Table 1 when they are trained on the Vimeo90K dataset.

	\subsection{Extrapolation Results}
	\label{ssec:experiment results}
	
	We evaluate the performance by measuring PSNR (Peak Signal-to-Noise Ratio) and SSIM (Structural Similarity) \cite{wang2004image} for all test datasets.
	Instead of using provided model, we train three interpolation models \cite{liu2017video, jiang2018super, niklaus2017video} from scratch to identify differences in extrapolation performance.
	We describe the performance differences of video extrapolation in Table \ref{main results}, where higher values of PSNR and SSIM show better extrapolation results. 
	In general, results with EIC loss roughly follow the performance of interpolation modules. For example, DVF and SepConv outperform SuperSlomo and baseline in all datasets and settings. If we choice interpolation module and hyper-parameter properly (\textit{e.g.} DVF and $\lambda=0.1$), the extrapolation performance (PSNR) is increased by 1.24 dB without changing network structures in the UCF101 dataset.
	
	Additionally, we train models with three hyper-parameter values: $\lambda=\{1, 0.1, 0.01\}$. In general, the models with $\lambda=0.1$ outperform the others. However, in KITTI dataset, some model with $\lambda=1$ shows better performance because when the videos have more complex motions, interpolation module can guide uncertain motion more effectively than the models with small $\lambda$ values.

	\subsection{Extrapolation Results on Long Future Frames}
	\label{ssec:longterm results}
	Another our purpose of designing EIC loss is to verify performance improvement in long future frame prediction. 
	To verify this, we conduct test for trained model to predict four future frames ($\hat{I}_{t}$, $\hat{I}_{t+1}$, $\hat{I}_{t+2}$, $\hat{I}_{t+3}$).
	When the number of input frames the model received is $k$, we can get $\hat{I}_{t}$ by equation~(\ref{extra_function}). In contrast, to obtain the next frames ($\hat{I}_{t+1}, \hat{I}_{t+2}, \hat{I}_{t+3}$), we do not have ideal inputs such as ($I_t, I_{t+1}, I_{t+2}$). 
	Therefore, in long future prediction, we use ($\hat{I}_{t}, \hat{I}_{t+1}, \hat{I}_{t+2}$) instead of ($I_t, I_{t+1}, I_{t+2}$) as the input frames. 
	This can cause performance decrement according to the number of predicted frames is increased for long future prediction due to the prediction error propagation.
	
	In Table \ref{longterm results}, we can see how the prediction performances are changed according to the future frames. 
	In general, due to the error propagation, PSNR and SSIM decrease when a predict frame is far from the input frames.
	Results show that our model can guarantee consistent prediction in long future frames from the observation that the performance gap (values in parenthesis) compared to the baseline model is increased.
	In Fig. \ref{fig:vis_results}, we describe these visual results and error maps in long future frames ($\hat{I}_{t+1}$,  $\hat{I}_{t+3}$), which show better visual results and lower error propagation (\textit{e.g.} artifacts) when prediction goes to the long futures.
	
	\section{Conclusion}
	\label{sec:conclusion}
	\vspace*{-4mm}
	
	In this work, we propose a novel Extrapolative-Interpolative Cycle (EIC) loss for video frame extrapolation. By adding our EIC loss, model can be learned at almost the same speed as baseline and produces improved performance without increasing memory usage. EIC loss can be applied to any combination of extrapolation and interpolation modules without modification of network structures. As shown in Section \ref{sec:experiments}, when our EIC loss is added, performance is hugely increased qualitatively and quantitatively. Since the interpolation guidance makes the uncertain prediction more stable, and better short future prediction quality mitigates the error propagation, long future frame prediction performance is also increased drastically.
	
	\indent \textbf{Acknowledgment.} This work was supported by Institute for Information \& communications Technology Promotion(IITP) grant funded by the Korea government(MSIP) (No.2016-0-00197, Development of the high-precision natural 3D view generation technology using smart-car multi sensors and deep learning)
	
	\newpage
	
	\bibliographystyle{IEEEbib}
	\small \bibliography{refs}
	
\end{document}